\pdfoutput=1

\documentclass[11pt]{article}

\usepackage{acl}

\usepackage{times}
\usepackage{latexsym}
\usepackage{todonotes}
\usepackage{url}

\usepackage[T1]{fontenc}

\usepackage[utf8]{inputenc}

\usepackage{microtype}
\usepackage{graphicx}
\usepackage{amsmath}

\title{(Un)solving Morphological Inflection: \\Lemma Overlap Artificially Inflates Models' Performance }

\author{Omer Goldman, \  David Guriel, \ Reut Tsarfaty \\
Bar-Ilan University \\
\texttt{\{omer.goldman,davidgu1312\}@gmail.com,reut.tsarfaty@biu.ac.il}}

\begin{document}
\maketitle
\begin{abstract}
In the domain of Morphology, Inflection is a fundamental and important task that gained a lot of traction in recent years, mostly via SIGMORPHON's shared-tasks.
With  average accuracy above 0.9 over the scores of all languages, the task is considered mostly solved using relatively generic neural seq2seq models, even with little data provided.
In this work, we propose to re-evaluate morphological inflection models by  employing harder train-test splits that will challenge the generalization capacity of the models. In particular, as opposed to the na{\"i}ve  split-by-form,  we propose a split-by-lemma method to challenge the performance on
existing benchmarks.
Our experiments with the three top-ranked systems on the   SIGMORPHON's 2020 shared-task show that the lemma-split presents an average drop of 30 percentage points in macro-average for the 90 languages included. The effect is most significant for low-resourced languages with a drop  as high as 95  points, but even high-resourced languages lose about 10 points on average. Our results clearly show that generalizing inflection to unseen lemmas is far from being solved, presenting a simple yet effective means to promote more sophisticated models.
\end{abstract}

\section{Introduction}
\label{sec:intro}
In  recent years, morphological  (re)inflection tasks in NLP have gained a lot of attention, most notably with the introduction of SIGMORPHON's shared tasks \cite{cotterell-etal-2016-sigmorphon, cotterell-etal-2017-conll, cotterell-etal-2018-conll, vylomova-etal-2020-sigmorphon} in tandem with the expansion of UniMorph \citep{mccarthy-etal-2020-unimorph}, a multi-lingual dataset of inflection tables. The shared-tasks sample data from UniMorph includes lists of triplets in the form of \textit{(lemma, features, form)} for many languages, and the shared-task organizers maintain standard splits for a fair system comparison.

The best-performing systems to-date in all inflection shared-tasks are neural sequence-to-sequence models used in many NLP tasks. An LSTM-based model won 2016's task \cite{kann-schutze-2016-med}, and a transformer came on top in 2020 \cite{canby-etal-2020-university}. In 2020's task the best model achieved exact-match accuracy that transcended 0.9 macro-averaged over up to 90 languages from various language families and types. %
This trend of high results recurred in works done on data collected independently as well (e.g. \citealp{malouf-2017-abstractive}, \citealp{silfverberg-hulden-2018-encoder}, inter alia).

Interestingly, the averaged results of 2020's shared-task include languages for which very little data was provided, sometimes as little as a couple of hundreds of examples. This has led to a view considering morphological inflection a relatively simple task that is essentially already {\em solved}, as reflected in the saturation of the results over the year and the declining submissions to the shared tasks.\footnote{The  shared task of 2021 had seen only two submissions \citep{pimentel-ryskina-etal-2021-sigmorphon}.} This also led the community to gravitate towards works attempting to solve the same (re)inflection tasks with little or no supervision \cite{mccarthy-etal-2019-sigmorphon, jin-etal-2020-unsupervised, goldman-tsarfaty-2021-minimal}.
    
However, before moving on we should ask ourselves whether morphological inflection is indeed {\em solved} or may the good performance %
be attributed to some artifacts in the data. This was shown to be true for many NLP tasks in which slight modifications of the data
can result in a more challenging dataset, e.g., the addition of unanswerable questions to question answering benchmarks  \citep{rajpurkar-etal-2018-know}, or the addition of expert-annotated minimal pairs to a variety of tasks \citep{gardner-etal-2020-evaluating}. A common modification is re-splitting the data such that the test set is more challenging and closer to the intended use of the models in the wild \citep{sogaard-etal-2021-need}. As the performance on morphological inflection models seems to have saturated on high scores, a similar rethinking of the data used is warranted.

In this work we propose to construct more difficult datasets for morphological (re)inflection by splitting them such that the test set will include no forms of lemmas appearing in the train set. This splitting method will allow assessing the models in a challenging scenario closer to their desired function in practice, where training data usually includes full inflection tables and learning to inflect the uncovered lemmas is the target.

We show, by re-splitting the data from task 0 of SIGMORPHON's 2020 shared-task, that the proposed split reveals a greater difficulty of morphological inflection. Retesting 3 of the 4 top-ranked systems of the shared-task on the new splits leads to a decrease of 30 points averaged over the systems for all 90 languages included in the shared-task. We further show that the effect is more prominent for low-resourced languages, where the drop can be as large as 95 points, though high-resourced languages may suffer from up to  a 10 points drop as well.
We conclude that in order to properly assess the performance  of (re)inflection models and to drive the field forward, the data and related splits should be carefully examined and  improved to provide a more challenging evaluation, more reflective of their real-world use.

\section{(Re)inflection and Memorization}
Inflection and reinflection are two of the most dominant tasks in  computational morphology. In the {\em inflection} task, the input is a lemma and a feature-bundle, and we aim to predict the respective inflected word-form. In {\em reinflection}, the input is an inflected word-form along with its features bundle, plus a feature-bundle without a form, and we aim to predict the respective inflected-form for the same lemma.
The {\em training} input in SIGMORPHON's shared-tasks is a random split of the available {\em (lemma,form,features)} triplets
such that no triplet occurring in the train-set occurs in the   test-set.\footnote{This is true for all SIGMORPHON's inflection shared tasks, save the paradigm completion task of 2017.}

In such a setting, models can  short-cut their way to better predictions in cases where forms from the same lemma appear in both the train and test data. This may allow models to memorize lemma-specific alternations that make morphological inflection a challenging task to begin with. Consider for example the notoriously unpredictable German plurality marking, where several allomorphs are associated with nouns with no clear rule governing the process. \textit{Kind}, for example, is pluralized with the suffix \textit{-er} resulting in \textit{Kinder} tagged as \textsc{nom;pl}. Assuming a model saw this example in the train set it is pretty easy to predict \textit{Kindern} for the same lemma with \textsc{dat;pl} features,\footnote{The addition of the dative marker \textit{-n} is very regular.} but without knowledge of the suffix used to pluralize \textit{Kind} the predictions \textit{Kinden} and \textit{Kinds} are just as likely.

\section{Related Work}
Many subfields of NLP and machine learning in general suggested {\em hard splits} as means to improve the probing of  models' ability to solve the underlying task,  and to make sure models do not simply employ loopholes in the data.

In the realm of sentence simplification,  \citet{narayan-etal-2017-split} suggested the \textsc{WebSplit} dataset,  where models are required to split and rephrase complex sentences associated with a meaning representation over a knowledge-base.
\citet{aharoni-goldberg-2018-split} found that some facts appeared in both train and test sets and provided a harder split denying models the ability to use memorized facts. 
\citet{aharoni-goldberg-2020-unsupervised} also suggested a general splitting method for machine translation such that the domains %
are as disjoint as possible.

In semantic parsing, \citet{finegan-dollak-etal-2018-improving} suggested a better split for parsing natural language questions to SQL queries by making sure that queries of the same template do not occur in both train and test, while \citet{lachmy-etal-2021-draw} split their \textsc{Hexagons} data such that any one visual pattern used for the task cannot appear in both train and test. Furthermore, \citet{loula-etal-2018-rearranging} adversarially split semantic parsing for navigation data to assess their models' capability to use compositionality.
In spoken language understanding \citet{arora-etal-2021-rethinking} designed a splitting method that will account for variation in both speaker identity and linguistic content.

In general, concerns regarding data splits and their undesired influence on model assessments led \citet{gorman-bedrick-2019-need} to advocate random splitting instead of standard ones. In reaction, \citet{sogaard-etal-2021-need} pointed to the flaws of random splits and suggested adversarial splits to challenge models further. Here we call for paying attention to the splits employed in evaluating morphological models, and improve on them.

\begin{table}[t]
\small{
\begin{center}
\begin{tabular}{lcccc}
\hline
 & \multicolumn{2}{c}{\textbf{Accuracy}} & \multicolumn{2}{c}{\textbf{Edit Distance}}\\
 \textbf{Split} & \textbf{Form} & \textbf{Lemma} & \textbf{Form} & \textbf{Lemma}\\
\hline
DeepSpin-02 & \textbf{0.90} & \textbf{0.76} & \textbf{0.23} & \textbf{0.58}\\ 
CULing & 0.88 & 0.63 & 0.29 & 1.02\\ 
Base trm-single & \textbf{0.90} & 0.53 & \textbf{0.23} & 1.32\\
Base LSTM & 0.85 & 0.39 & 0.34 & 1.79\\ 
\hline
\textbf{Average} & 0.88 & 0.58 & 0.27 & 1.18\\
\hline
\end{tabular}
\end{center}
}
\caption{Exact-match accuracy and edit-distance for our baseline and 3 of the 4 top-ranked systems of SIGMORPHON's 2020 shared-task, all reported on the original split of the shared-task (form split) and on our harder lemma split. Best system per column is in \textbf{bold}.}
\label{tab:main_res}
\end{table}

\begin{table}[t]
\small{
\begin{center}
\begin{tabular}{lcccc}
\hline
 & \multicolumn{2}{c}{\textbf{Accuracy}} \\
\textbf{Split} & \textbf{Form} & \textbf{Lemma} \\
\hline
Afro-Asiatic & 0.93 (0.95)$_\textsc{t}$ & 0.51 (0.80)$_\textsc{d}$ \\
Austronesian & 0.78 (0.82)$_\textsc{t}$ & 0.45 (0.70)$_\textsc{d}$ \\ 
Germanic & 0.86 (0.88)$_\textsc{d}$ & 0.63 (0.74)$_\textsc{d}$ \\ 
Indo-Iranian & 0.93 (0.97)$_\textsc{d}$ & 0.55 (0.86)$_\textsc{d}$ \\ 
Niger-Congo & 0.95 (0.98)$_\textsc{t}$ & 0.56 (0.90)$_\textsc{d}$ \\ 
Oto-Manguean & 0.84 (0.86)$_\textsc{t}$ & 0.53 (0.60)$_\textsc{d}$ \\
Romance  & 0.97 (0.99)$_\textsc{t}$ & 0.69 (0.86)$_\textsc{d}$ \\
Turkic & 0.95 (0.96)$_\textsc{t}$ & 0.64 (0.89)$_\textsc{d}$ \\
Uralic & 0.88 (0.90)$_\textsc{c}$ & 0.65 (0.72)$_\textsc{d}$ \\\hline
\end{tabular}
\end{center}
}
\caption{Aggregated results for the various language families. We provide the performance averaged across all systems, and in parenthesis the performance of the best system per family. The best system is identifiable in subscript: \textsc{c} - CULing, \textsc{t} - Base trm-single, \textsc{d} - DeepSpin-02. We include here only families with at least 3 languages in the data.}
\label{tab:res-by-family}
\end{table}

\section{Experiments}
In order to better assess the difficulty of morphological inflection, we compare the performances of 3 of the top-ranked system at task 0 (inflection) of SIGMORPHON's 2020 shared-task. We examined each system on both the the standard (form) split and the novel (lemma) split.

When re-splitting,\footnote{The split was done randomly as is  standard in SIGMORPHON tasks, although frequency-based sampling is also conceivable and is sometimes used, as in \citet{cotterell-etal-2018-conll}.} we kept the same proportions of the form-split data, i.e. we split the inflection tables 70\%, 10\% and 20\% for the train, development and test set. 
In terms of examples the proportions may vary as not all tables are of equal size. In practice, the averaged train set size in examples terms was only 3.5\% smaller in the lemma-split data, on average.\footnote{The newly-split data is available at \url{https://github.com/OnlpLab/LemmaSplitting}.}

\subsection{The Languages}
SIGMORHPON's 2020 shared-task includes datasets for 90 typologically and genealogically diverse languages from 14 language families. %
The languages are varied along almost any typological dimension, from fusional to agglutinative, small inflection tables to vast ones. They include mostly prefixing and mostly suffixing languages with representation of infixing and circumfixing as well. The languages vary also in use, including widely-used languages such as English and Hindi and moribund or extinct languages like Dakota and Middle High German.\footnote{The full list with the originally released data are at \url{https://github.com/sigmorphon2020/task0-data}.}

\subsection{The Models}
We tested the effects of lemma-splitting on our own LSTM-based model as well as 3 of the 4 top-ranked systems in the shared task.\footnote{The best performing system, UIUC \citep{canby-etal-2020-university}, did not have a publicly available implementation.}

\paragraph{Base LSTM}
We implemented a character-based sequence-to-sequence model which consists of a 1-layer bi-directional LSTM Encoder and a 1-layer unidirectional LSTM Decoder with a global soft attention layer \citep{bahdanau-etal-2014-neural}. Our model was trained for 50 epochs with no model selection.\footnote{The code is available at \url{https://github.com/OnlpLab/LemmaSplitting}.} 

\paragraph{Base trm-single}
The shared-task's organizers supplied various baselines, some based on a transformer architecture that was adapted for character-level tasks \citep{wu-etal-2021-applying}.\footnote{The code is available at \url{https://github.com/shijie-wu/neural-transducer}.} All baseline models include 4 encoder and 4 decoder layers, consisting of a multi-head self-attention layer and 2 feed-forward layers, equipped with a skip-connection. In every decoder layer a multi-head attention layer attends to the encoder's outputs. The network was trained for 4,000 warm-up steps and up to 20,000 more steps, each step over a batch of size 400. The model was examined with and without augmented data, trained separately on each language or each language family.
One of the baseline setups, training a model per language without augmented data, made it to the top 4 systems and we  include it here.

\paragraph{DeepSpin}
\citet{peters-martins-2020-one} submitted a recurrent neural network -- dubbed DeepSpin\nobreakdash-02.\footnote{The code is available at \url{https://github.com/deep-spin/sigmorphon-seq2seq}.} The system is composed of 2 bi-directional LSTM encoders with bi-linear gated Attention \citep{Luong-etal-2015-effective}, one for the lemma characters and one for the features characters, and a unidirectional LSTM Decoder for generating the outputs.
The innovation in the architecture is the use of sparsemax \citep{martins-astudillo-2016-from} instead of softmax in the attention layer.\footnote{The system submitted as DeepSpin-01 uses 1.5-entmax \citep{peters-martins-2019-ist} rather than sparsemax. Both systems perform highly similarly, hence we do not detail results for both.}

\paragraph{CULing}
\citet{liu-hulden-2020-leveraging}'s system is also based on the transformer architecture, with hyper-parameters very similar to \textit{base trm-single}.\footnote{The code is available at \url{https://github.com/LINGuistLIU/principal_parts_for_inflection}.}
Their innovation is in restructuring the data such that the model learns to inflect from any given cell in the inflection table rather than solely from the lemma.

\subsection{Results}
Table \ref{tab:main_res} summarizes our main results. We clearly see a drop in the performance for all systems, with an average of 30 points. The table also shows that splitting the data according to lemmas allows discerning between  systems that appear to perform quite similarly on the form-split data. The best system on the lemma-split data, DeepSpin-02, outperforms the second-ranked CULing system by about 13 points with both baseline systems performing significantly worse. The results in terms of averaged edit distance show the same trends.

DeepSpin-02 emerges victorious also in Table \ref{tab:res-by-family}, where results are broken down by language family. The table shows that DeepSpin-02 is the best performer over all language families when data is lemma-split, in contrast to the mixed picture over the form-split data.

\begin{figure}[t]
\includegraphics[width=1\columnwidth]{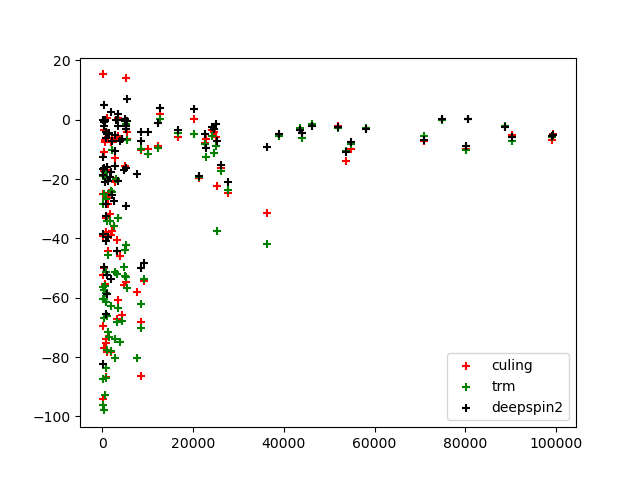}
\caption{Performance drop for the various systems when moving from form to lemma split as a function of the size of the train data. The effect is clearly more significant for lower-resourced languages.}
\label{fig:size-break-down}
\end{figure}

\begin{figure}[t]
\includegraphics[width=1\columnwidth]{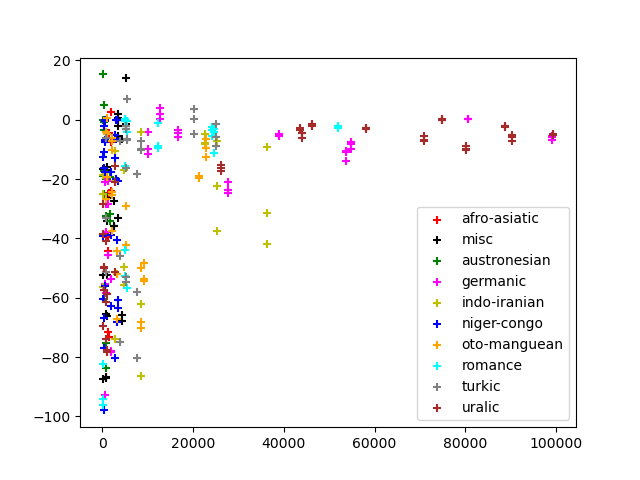}
\caption{Performance drop for the various language families when moving from form to lemma split as a function of the size of the train data. We include here only families with at least 3 languages in the data, the rest are classified under \textit{misc}.}
\label{fig:size-break-down-by-family}
\end{figure}

The average performance per language family seems to be controlled by training data availability.
For example, Germanic languages show average drop of 23 points, while for Niger-Congo languages the drop is 39 points on average.

In order to further examine the relation between the amount of training data and drop in performance we plotted in Figure~\ref{fig:size-break-down} the drop per system and per language against the size of the available train data, color-coded to indicate systems. It shows that the major drops in performance that contributed the most to the overall gap between the splits are in those low-resourced language. Remarkably, for some systems and languages the drop can be as high as 95 points. On the other hand, on high-resourced languages with 40,000 training examples or more,  all systems didn't lose much.
The analysis also shows the advantage of DeepSpin-02 in the lower-resourced settings that made it the best performer overall.

When color-coding the same broken-down data for linguistic family membership rather than system, as we do in Figure~\ref{fig:size-break-down-by-family}, it becomes clear that there is no evidence for specific families being easier for inflection when little data is provided. The figure does show the remarkable discrepancy in annotation effort, as the high-resourced languages mostly belong to 2 families: Germanic and Uralic.

\section{Discussion}

We proposed a method for splitting morphological datasets such that there is no lemma overlap between the splits. On the re-split of SIGMORPHON's 2020 shared-task data, we showed that all top-ranked systems suffer significant drops in performance. 
The new split examines models' generalization abilities in conditions more similar to their desired usage in the wild and allows better discerning between the systems in order to point to more promising directions for future research --- more so than the original form-split data on which all systems fared similarly. The new splitting method is likely to lead to more sophisticated modeling, for instance, in the spirit of the model proposed by \citet{liu-hulden-2021-can}.
The suggested move to a harder split is not unlike many other NLP tasks, in which challenging splits are suggested to drive the field forward. We thus call for morphological studies to carefully attend to the data used and expose the actual difficulties in modelling morphology, in future research and future shared tasks. %

\section*{Acknowledgements}

This research was funded by the European Research Council under the European Union’s Horizon 2020 research and innovation programme, (grant agreement No.\ 677352)  and by a research grant from the ministry of Science and Technology (MOST) of the Israeli Government, for which we are grateful.

\bibliography{anthology,custom}

\begin{thebibliography}{32}
\expandafter\ifx\csname natexlab\endcsname\relax\def\natexlab#1{#1}\fi

\bibitem[{Aharoni and Goldberg(2018)}]{aharoni-goldberg-2018-split}
Roee Aharoni and Yoav Goldberg. 2018.
\newblock \href {https://doi.org/10.18653/v1/P18-2114} {Split and rephrase:
  Better evaluation and stronger baselines}.
\newblock In \emph{Proceedings of the 56th Annual Meeting of the Association
  for Computational Linguistics (Volume 2: Short Papers)}, pages 719--724,
  Melbourne, Australia. Association for Computational Linguistics.

\bibitem[{Aharoni and Goldberg(2020)}]{aharoni-goldberg-2020-unsupervised}
Roee Aharoni and Yoav Goldberg. 2020.
\newblock \href {https://doi.org/10.18653/v1/2020.acl-main.692} {Unsupervised
  domain clusters in pretrained language models}.
\newblock In \emph{Proceedings of the 58th Annual Meeting of the Association
  for Computational Linguistics}, pages 7747--7763, Online. Association for
  Computational Linguistics.

\bibitem[{Arora et~al.(2021)Arora, Ostapenko, Viswanathan, Dalmia, Metze,
  Watanabe, and Black}]{arora-etal-2021-rethinking}
Siddhant Arora, Alissa Ostapenko, Vijay Viswanathan, Siddharth Dalmia, Florian
  Metze, Shinji Watanabe, and Alan~W Black. 2021.
\newblock \href {http://arxiv.org/abs/2106.15065} {Rethinking end-to-end
  evaluation of decomposable tasks: A case study on spoken language
  understanding}.

\bibitem[{Bahdanau et~al.(2014)Bahdanau, Cho, and
  Bengio}]{bahdanau-etal-2014-neural}
Dzmitry Bahdanau, Kyunghyun Cho, and Yoshua Bengio. 2014.
\newblock \href {http://arxiv.org/abs/1409.0473} {Neural machine translation by
  jointly learning to align and translate}.
\newblock Cite arxiv:1409.0473Comment: Accepted at ICLR 2015 as oral
  presentation.

\bibitem[{Canby et~al.(2020)Canby, Karipbayeva, Lunt, Mozaffari, Yoder, and
  Hockenmaier}]{canby-etal-2020-university}
Marc Canby, Aidana Karipbayeva, Bryan Lunt, Sahand Mozaffari, Charlotte Yoder,
  and Julia Hockenmaier. 2020.
\newblock \href {https://doi.org/10.18653/v1/2020.sigmorphon-1.15}
  {{U}niversity of {I}llinois submission to the {SIGMORPHON} 2020 shared task
  0: Typologically diverse morphological inflection}.
\newblock In \emph{Proceedings of the 17th SIGMORPHON Workshop on Computational
  Research in Phonetics, Phonology, and Morphology}, pages 137--145, Online.
  Association for Computational Linguistics.

\bibitem[{Cotterell et~al.(2018)Cotterell, Kirov, Sylak-Glassman, Walther,
  Vylomova, McCarthy, Kann, Mielke, Nicolai, Silfverberg, Yarowsky, Eisner, and
  Hulden}]{cotterell-etal-2018-conll}
Ryan Cotterell, Christo Kirov, John Sylak-Glassman, G{\'e}raldine Walther,
  Ekaterina Vylomova, Arya~D. McCarthy, Katharina Kann, Sabrina~J. Mielke,
  Garrett Nicolai, Miikka Silfverberg, David Yarowsky, Jason Eisner, and Mans
  Hulden. 2018.
\newblock \href {https://doi.org/10.18653/v1/K18-3001} {The
  {C}o{NLL}{--}{SIGMORPHON} 2018 shared task: Universal morphological
  reinflection}.
\newblock In \emph{Proceedings of the {C}o{NLL}{--}{SIGMORPHON} 2018 Shared
  Task: Universal Morphological Reinflection}, pages 1--27, Brussels.
  Association for Computational Linguistics.

\bibitem[{Cotterell et~al.(2017)Cotterell, Kirov, Sylak-Glassman, Walther,
  Vylomova, Xia, Faruqui, K{\"u}bler, Yarowsky, Eisner, and
  Hulden}]{cotterell-etal-2017-conll}
Ryan Cotterell, Christo Kirov, John Sylak-Glassman, G{\'e}raldine Walther,
  Ekaterina Vylomova, Patrick Xia, Manaal Faruqui, Sandra K{\"u}bler, David
  Yarowsky, Jason Eisner, and Mans Hulden. 2017.
\newblock \href {https://doi.org/10.18653/v1/K17-2001} {{C}o{NLL}-{SIGMORPHON}
  2017 shared task: Universal morphological reinflection in 52 languages}.
\newblock In \emph{Proceedings of the {C}o{NLL} {SIGMORPHON} 2017 Shared Task:
  Universal Morphological Reinflection}, pages 1--30, Vancouver. Association
  for Computational Linguistics.

\bibitem[{Cotterell et~al.(2016)Cotterell, Kirov, Sylak-Glassman, Yarowsky,
  Eisner, and Hulden}]{cotterell-etal-2016-sigmorphon}
Ryan Cotterell, Christo Kirov, John Sylak-Glassman, David Yarowsky, Jason
  Eisner, and Mans Hulden. 2016.
\newblock \href {https://doi.org/10.18653/v1/W16-2002} {The {SIGMORPHON} 2016
  shared {T}ask{---}{M}orphological reinflection}.
\newblock In \emph{Proceedings of the 14th {SIGMORPHON} Workshop on
  Computational Research in Phonetics, Phonology, and Morphology}, pages
  10--22, Berlin, Germany. Association for Computational Linguistics.

\bibitem[{Finegan-Dollak et~al.(2018)Finegan-Dollak, Kummerfeld, Zhang,
  Ramanathan, Sadasivam, Zhang, and Radev}]{finegan-dollak-etal-2018-improving}
Catherine Finegan-Dollak, Jonathan~K. Kummerfeld, Li~Zhang, Karthik Ramanathan,
  Sesh Sadasivam, Rui Zhang, and Dragomir Radev. 2018.
\newblock \href {https://doi.org/10.18653/v1/P18-1033} {Improving text-to-{SQL}
  evaluation methodology}.
\newblock In \emph{Proceedings of the 56th Annual Meeting of the Association
  for Computational Linguistics (Volume 1: Long Papers)}, pages 351--360,
  Melbourne, Australia. Association for Computational Linguistics.

\bibitem[{Gardner et~al.(2020)Gardner, Artzi, Basmov, Berant, Bogin, Chen,
  Dasigi, Dua, Elazar, Gottumukkala, Gupta, Hajishirzi, Ilharco, Khashabi, Lin,
  Liu, Liu, Mulcaire, Ning, Singh, Smith, Subramanian, Tsarfaty, Wallace,
  Zhang, and Zhou}]{gardner-etal-2020-evaluating}
Matt Gardner, Yoav Artzi, Victoria Basmov, Jonathan Berant, Ben Bogin, Sihao
  Chen, Pradeep Dasigi, Dheeru Dua, Yanai Elazar, Ananth Gottumukkala, Nitish
  Gupta, Hannaneh Hajishirzi, Gabriel Ilharco, Daniel Khashabi, Kevin Lin,
  Jiangming Liu, Nelson~F. Liu, Phoebe Mulcaire, Qiang Ning, Sameer Singh,
  Noah~A. Smith, Sanjay Subramanian, Reut Tsarfaty, Eric Wallace, Ally Zhang,
  and Ben Zhou. 2020.
\newblock \href {https://doi.org/10.18653/v1/2020.findings-emnlp.117}
  {Evaluating models{'} local decision boundaries via contrast sets}.
\newblock In \emph{Findings of the Association for Computational Linguistics:
  EMNLP 2020}, pages 1307--1323, Online. Association for Computational
  Linguistics.

\bibitem[{Goldman and Tsarfaty(2021)}]{goldman-tsarfaty-2021-minimal}
Omer Goldman and Reut Tsarfaty. 2021.
\newblock \href {http://arxiv.org/abs/2104.08512} {Minimal supervision for
  morphological inflection}.

\bibitem[{Gorman and Bedrick(2019)}]{gorman-bedrick-2019-need}
Kyle Gorman and Steven Bedrick. 2019.
\newblock \href {https://doi.org/10.18653/v1/P19-1267} {We need to talk about
  standard splits}.
\newblock In \emph{Proceedings of the 57th Annual Meeting of the Association
  for Computational Linguistics}, pages 2786--2791, Florence, Italy.
  Association for Computational Linguistics.

\bibitem[{Jin et~al.(2020)Jin, Cai, Peng, Xia, McCarthy, and
  Kann}]{jin-etal-2020-unsupervised}
Huiming Jin, Liwei Cai, Yihui Peng, Chen Xia, Arya McCarthy, and Katharina
  Kann. 2020.
\newblock \href {https://doi.org/10.18653/v1/2020.acl-main.598} {Unsupervised
  morphological paradigm completion}.
\newblock In \emph{Proceedings of the 58th Annual Meeting of the Association
  for Computational Linguistics}, pages 6696--6707, Online. Association for
  Computational Linguistics.

\bibitem[{Kann and Sch{\"u}tze(2016)}]{kann-schutze-2016-med}
Katharina Kann and Hinrich Sch{\"u}tze. 2016.
\newblock \href {https://doi.org/10.18653/v1/W16-2010} {{MED}: The {LMU} system
  for the {SIGMORPHON} 2016 shared task on morphological reinflection}.
\newblock In \emph{Proceedings of the 14th {SIGMORPHON} Workshop on
  Computational Research in Phonetics, Phonology, and Morphology}, pages
  62--70, Berlin, Germany. Association for Computational Linguistics.

\bibitem[{Lachmy et~al.(2021)Lachmy, Pyatkin, and
  Tsarfaty}]{lachmy-etal-2021-draw}
Royi Lachmy, Valentina Pyatkin, and Reut Tsarfaty. 2021.
\newblock \href {http://arxiv.org/abs/2106.14321} {Draw me a flower: Grounding
  formal abstract structures stated in informal natural language}.

\bibitem[{Liu and Hulden(2020)}]{liu-hulden-2020-leveraging}
Ling Liu and Mans Hulden. 2020.
\newblock \href {https://doi.org/10.18653/v1/2020.sigmorphon-1.17} {Leveraging
  principal parts for morphological inflection}.
\newblock In \emph{Proceedings of the 17th SIGMORPHON Workshop on Computational
  Research in Phonetics, Phonology, and Morphology}, pages 153--161, Online.
  Association for Computational Linguistics.

\bibitem[{Liu and Hulden(2021)}]{liu-hulden-2021-can}
Ling Liu and Mans Hulden. 2021.
\newblock \href {http://arxiv.org/abs/2104.06483} {Can a transformer pass the
  wug test? tuning copying bias in neural morphological inflection models}.
\newblock \emph{CoRR}, abs/2104.06483.

\bibitem[{Loula et~al.(2018)Loula, Baroni, and
  Lake}]{loula-etal-2018-rearranging}
Jo{\~a}o Loula, Marco Baroni, and Brenden Lake. 2018.
\newblock \href {https://doi.org/10.18653/v1/W18-5413} {Rearranging the
  familiar: Testing compositional generalization in recurrent networks}.
\newblock In \emph{Proceedings of the 2018 {EMNLP} Workshop {B}lackbox{NLP}:
  Analyzing and Interpreting Neural Networks for {NLP}}, pages 108--114,
  Brussels, Belgium. Association for Computational Linguistics.

\bibitem[{Luong et~al.(2015)Luong, Pham, and
  Manning}]{Luong-etal-2015-effective}
Thang Luong, Hieu Pham, and Christopher~D. Manning. 2015.
\newblock \href {https://doi.org/10.18653/v1/D15-1166} {Effective approaches to
  attention-based neural machine translation}.
\newblock In \emph{Proceedings of the 2015 Conference on Empirical Methods in
  Natural Language Processing}, pages 1412--1421, Lisbon, Portugal. Association
  for Computational Linguistics.

\bibitem[{Malouf(2017)}]{malouf-2017-abstractive}
Robert Malouf. 2017.
\newblock Abstractive morphological learning with a recurrent neural network.
\newblock \emph{Morphology}, 27(4):431--458.

\bibitem[{Martins and Astudillo(2016)}]{martins-astudillo-2016-from}
Andr\'{e} F.~T. Martins and Ram\'{o}n~F. Astudillo. 2016.
\newblock \href {https://arxiv.org/pdf/1602.02068.pdf} {From softmax to
  sparsemax: A sparse model of attention and multi-label classification}.
\newblock In \emph{Proceedings of the 33rd International Conference on
  International Conference on Machine Learning - Volume 48}, ICML'16, page
  1614–1623. JMLR.org.

\bibitem[{McCarthy et~al.(2020)McCarthy, Kirov, Grella, Nidhi, Xia, Gorman,
  Vylomova, Mielke, Nicolai, Silfverberg, Arkhangelskiy, Krizhanovsky,
  Krizhanovsky, Klyachko, Sorokin, Mansfield, Ern{\v{s}}treits, Pinter, Jacobs,
  Cotterell, Hulden, and Yarowsky}]{mccarthy-etal-2020-unimorph}
Arya~D. McCarthy, Christo Kirov, Matteo Grella, Amrit Nidhi, Patrick Xia, Kyle
  Gorman, Ekaterina Vylomova, Sabrina~J. Mielke, Garrett Nicolai, Miikka
  Silfverberg, Timofey Arkhangelskiy, Nataly Krizhanovsky, Andrew Krizhanovsky,
  Elena Klyachko, Alexey Sorokin, John Mansfield, Valts Ern{\v{s}}treits, Yuval
  Pinter, Cassandra~L. Jacobs, Ryan Cotterell, Mans Hulden, and David Yarowsky.
  2020.
\newblock \href {https://aclanthology.org/2020.lrec-1.483} {{U}ni{M}orph 3.0:
  {U}niversal {M}orphology}.
\newblock In \emph{Proceedings of the 12th Language Resources and Evaluation
  Conference}, pages 3922--3931, Marseille, France. European Language Resources
  Association.

\bibitem[{McCarthy et~al.(2019)McCarthy, Vylomova, Wu, Malaviya, Wolf-Sonkin,
  Nicolai, Kirov, Silfverberg, Mielke, Heinz, Cotterell, and
  Hulden}]{mccarthy-etal-2019-sigmorphon}
Arya~D. McCarthy, Ekaterina Vylomova, Shijie Wu, Chaitanya Malaviya, Lawrence
  Wolf-Sonkin, Garrett Nicolai, Christo Kirov, Miikka Silfverberg, Sabrina~J.
  Mielke, Jeffrey Heinz, Ryan Cotterell, and Mans Hulden. 2019.
\newblock \href {https://doi.org/10.18653/v1/W19-4226} {The {SIGMORPHON} 2019
  shared task: Morphological analysis in context and cross-lingual transfer for
  inflection}.
\newblock In \emph{Proceedings of the 16th Workshop on Computational Research
  in Phonetics, Phonology, and Morphology}, pages 229--244, Florence, Italy.
  Association for Computational Linguistics.

\bibitem[{Narayan et~al.(2017)Narayan, Gardent, Cohen, and
  Shimorina}]{narayan-etal-2017-split}
Shashi Narayan, Claire Gardent, Shay~B. Cohen, and Anastasia Shimorina. 2017.
\newblock \href {https://doi.org/10.18653/v1/D17-1064} {Split and rephrase}.
\newblock In \emph{Proceedings of the 2017 Conference on Empirical Methods in
  Natural Language Processing}, pages 606--616, Copenhagen, Denmark.
  Association for Computational Linguistics.

\bibitem[{Peters and Martins(2019)}]{peters-martins-2019-ist}
Ben Peters and Andr{\'e} F.~T. Martins. 2019.
\newblock \href {https://doi.org/10.18653/v1/W19-4207} {{IT}{--}{IST} at the
  {SIGMORPHON} 2019 shared task: Sparse two-headed models for inflection}.
\newblock In \emph{Proceedings of the 16th Workshop on Computational Research
  in Phonetics, Phonology, and Morphology}, pages 50--56, Florence, Italy.
  Association for Computational Linguistics.

\bibitem[{Peters and Martins(2020)}]{peters-martins-2020-one}
Ben Peters and Andr{\'e} F.~T. Martins. 2020.
\newblock \href {https://doi.org/10.18653/v1/2020.sigmorphon-1.4}
  {One-size-fits-all multilingual models}.
\newblock In \emph{Proceedings of the 17th SIGMORPHON Workshop on Computational
  Research in Phonetics, Phonology, and Morphology}, pages 63--69, Online.
  Association for Computational Linguistics.

\bibitem[{Pimentel et~al.(2021)Pimentel, Ryskina, Mielke, Wu, Chodroff,
  Leonard, Nicolai, Ghanggo~Ate, Khalifa, Habash, El-Khaissi, Goldman, Gasser,
  Lane, Coler, Oncevay, Montoya~Samame, Silva~Villegas, Ek, Bernardy,
  Shcherbakov, Bayyr-ool, Sheifer, Ganieva, Plugaryov, Klyachko, Salehi,
  Krizhanovsky, Krizhanovsky, Vania, Ivanova, Salchak, Straughn, Liu,
  Washington, Ataman, Kiera{\'s}, Woli{\'n}ski, Suhardijanto, Stoehr, Nuriah,
  Ratan, Tyers, Ponti, Aiton, Hatcher, Prud'hommeaux, Kumar, Hulden, Barta,
  Lakatos, Szolnok, {\'A}cs, Raj, Yarowsky, Cotterell, Ambridge, and
  Vylomova}]{pimentel-ryskina-etal-2021-sigmorphon}
Tiago Pimentel, Maria Ryskina, Sabrina~J. Mielke, Shijie Wu, Eleanor Chodroff,
  Brian Leonard, Garrett Nicolai, Yustinus Ghanggo~Ate, Salam Khalifa, Nizar
  Habash, Charbel El-Khaissi, Omer Goldman, Michael Gasser, William Lane, Matt
  Coler, Arturo Oncevay, Jaime~Rafael Montoya~Samame, Gema~Celeste
  Silva~Villegas, Adam Ek, Jean-Philippe Bernardy, Andrey Shcherbakov, Aziyana
  Bayyr-ool, Karina Sheifer, Sofya Ganieva, Matvey Plugaryov, Elena Klyachko,
  Ali Salehi, Andrew Krizhanovsky, Natalia Krizhanovsky, Clara Vania, Sardana
  Ivanova, Aelita Salchak, Christopher Straughn, Zoey Liu, Jonathan~North
  Washington, Duygu Ataman, Witold Kiera{\'s}, Marcin Woli{\'n}ski, Totok
  Suhardijanto, Niklas Stoehr, Zahroh Nuriah, Shyam Ratan, Francis~M. Tyers,
  Edoardo~M. Ponti, Grant Aiton, Richard~J. Hatcher, Emily Prud'hommeaux,
  Ritesh Kumar, Mans Hulden, Botond Barta, Dorina Lakatos, G{\'a}bor Szolnok,
  Judit {\'A}cs, Mohit Raj, David Yarowsky, Ryan Cotterell, Ben Ambridge, and
  Ekaterina Vylomova. 2021.
\newblock \href {https://doi.org/10.18653/v1/2021.sigmorphon-1.25} {Sigmorphon
  2021 shared task on morphological reinflection: Generalization across
  languages}.
\newblock In \emph{Proceedings of the 18th SIGMORPHON Workshop on Computational
  Research in Phonetics, Phonology, and Morphology}, pages 229--259, Online.
  Association for Computational Linguistics.

\bibitem[{Rajpurkar et~al.(2018)Rajpurkar, Jia, and
  Liang}]{rajpurkar-etal-2018-know}
Pranav Rajpurkar, Robin Jia, and Percy Liang. 2018.
\newblock \href {https://doi.org/10.18653/v1/P18-2124} {Know what you don{'}t
  know: Unanswerable questions for {SQ}u{AD}}.
\newblock In \emph{Proceedings of the 56th Annual Meeting of the Association
  for Computational Linguistics (Volume 2: Short Papers)}, pages 784--789,
  Melbourne, Australia. Association for Computational Linguistics.

\bibitem[{Silfverberg and Hulden(2018)}]{silfverberg-hulden-2018-encoder}
Miikka Silfverberg and Mans Hulden. 2018.
\newblock \href {https://doi.org/10.18653/v1/D18-1315} {An encoder-decoder
  approach to the paradigm cell filling problem}.
\newblock In \emph{Proceedings of the 2018 Conference on Empirical Methods in
  Natural Language Processing}, pages 2883--2889, Brussels, Belgium.
  Association for Computational Linguistics.

\bibitem[{S{\o}gaard et~al.(2021)S{\o}gaard, Ebert, Bastings, and
  Filippova}]{sogaard-etal-2021-need}
Anders S{\o}gaard, Sebastian Ebert, Jasmijn Bastings, and Katja Filippova.
  2021.
\newblock \href {https://aclanthology.org/2021.eacl-main.156} {We need to talk
  about random splits}.
\newblock In \emph{Proceedings of the 16th Conference of the European Chapter
  of the Association for Computational Linguistics: Main Volume}, pages
  1823--1832, Online. Association for Computational Linguistics.

\bibitem[{Vylomova et~al.(2020)Vylomova, White, Salesky, Mielke, Wu, Ponti,
  Hall~Maudslay, Zmigrod, Valvoda, Toldova, Tyers, Klyachko, Yegorov,
  Krizhanovsky, Czarnowska, Nikkarinen, Krizhanovsky, Pimentel,
  Torroba~Hennigen, Kirov, Nicolai, Williams, Anastasopoulos, Cruz, Chodroff,
  Cotterell, Silfverberg, and Hulden}]{vylomova-etal-2020-sigmorphon}
Ekaterina Vylomova, Jennifer White, Elizabeth Salesky, Sabrina~J. Mielke,
  Shijie Wu, Edoardo~Maria Ponti, Rowan Hall~Maudslay, Ran Zmigrod, Josef
  Valvoda, Svetlana Toldova, Francis Tyers, Elena Klyachko, Ilya Yegorov,
  Natalia Krizhanovsky, Paula Czarnowska, Irene Nikkarinen, Andrew
  Krizhanovsky, Tiago Pimentel, Lucas Torroba~Hennigen, Christo Kirov, Garrett
  Nicolai, Adina Williams, Antonios Anastasopoulos, Hilaria Cruz, Eleanor
  Chodroff, Ryan Cotterell, Miikka Silfverberg, and Mans Hulden. 2020.
\newblock \href {https://doi.org/10.18653/v1/2020.sigmorphon-1.1} {{SIGMORPHON}
  2020 shared task 0: Typologically diverse morphological inflection}.
\newblock In \emph{Proceedings of the 17th SIGMORPHON Workshop on Computational
  Research in Phonetics, Phonology, and Morphology}, pages 1--39, Online.
  Association for Computational Linguistics.

\bibitem[{Wu et~al.(2021)Wu, Cotterell, and Hulden}]{wu-etal-2021-applying}
Shijie Wu, Ryan Cotterell, and Mans Hulden. 2021.
\newblock \href {https://aclanthology.org/2021.eacl-main.163} {Applying the
  transformer to character-level transduction}.
\newblock In \emph{Proceedings of the 16th Conference of the European Chapter
  of the Association for Computational Linguistics: Main Volume}, pages
  1901--1907, Online. Association for Computational Linguistics.

\end{thebibliography}
\bibliographystyle{acl_natbib}

\end{document}